\def\year{2020}\relax
\documentclass[letterpaper]{article} % DO NOT CHANGE THIS
\usepackage{aaai20}  % DO NOT CHANGE THIS
\usepackage{times}  % DO NOT CHANGE THIS
\usepackage{helvet} % DO NOT CHANGE THIS
\usepackage{courier}  % DO NOT CHANGE THIS
\usepackage[hyphens]{url}  % DO NOT CHANGE THIS
\usepackage{graphicx} % DO NOT CHANGE THIS
\usepackage{graphicx}  % DO NOT CHANGE THIS
\usepackage{booktabs}
\usepackage{multirow,bigdelim}
\usepackage{verbatim}
\usepackage{amsmath}
\usepackage{bbm}
\usepackage{amssymb}
\title{Mask \& Focus: Conversation Modelling by Learning Concepts}
\author{
    Gaurav Pandey, Dinesh Raghu and Sachindra Joshi\\
    IBM Research, New Delhi, India\\
    \textrm{\{gpandey1, diraghu1, jsachind\}@in.ibm.com}
}
\begin{document}

\maketitle

\begin{abstract}     
Sequence to sequence models attempt to capture the correlation between all the words in the input and output sequences. While this is quite useful for machine translation where the correlation among the words is indeed quite strong, it becomes problematic for conversation modelling where the correlation is often at a much abstract level. In contrast, humans tend to focus on the essential concepts discussed in the conversation context and generate responses accordingly. In this paper, we attempt to mimic this response generating mechanism by learning the essential concepts in the context and response in an unsupervised manner. The proposed model, referred to as Mask \& Focus maps the input context to a sequence of concepts which are then used to generate the response concepts. Together, the context and the response concepts generate the final response. In order to learn context concepts from the training data automatically, we \emph{mask} words in the input and observe the effect of masking on response generation. We train our model to learn those response concepts that have high mutual information with respect to the context concepts, thereby guiding the model to \emph{focus} on the context concepts. Mask \& Focus achieves significant improvement over the existing baselines in several established metrics for dialogues.
%In our experimental results on Ubuntu Dialogue Corpus and Technical Support Dataset, we observe a significant improvement in several established metrics for dialogues.
\end{abstract}

\section{Introduction}

Standard neural conversation modelling~\cite{vinyals2015neural,serban2016building} follows a sequence-to-sequence framework for learning the relationship between sequence of words in the conversation-context and the next response. Attempts have been made to incorporate abstract information in the form of latent variables to introduce variability in response generation~\cite{zhao2017learning,serban2017hierarchical}. 
In these architectures, all words in the conversation-context and the next response are treated to be equally important. Such an assumption works well for machine translation as most words/phrases in the source language have to be encoded and translated to the target language. However, for modelling conversations, the encoder is expected to understand the conversation context with a higher level of abstraction to generate the next response.
%But these models hope that given enough examples, the model will learn to discern the parts in the context that are essential for responding while ignoring other unnecessary information. 
\begin{figure}
    \centering
    \includegraphics[scale=0.35]{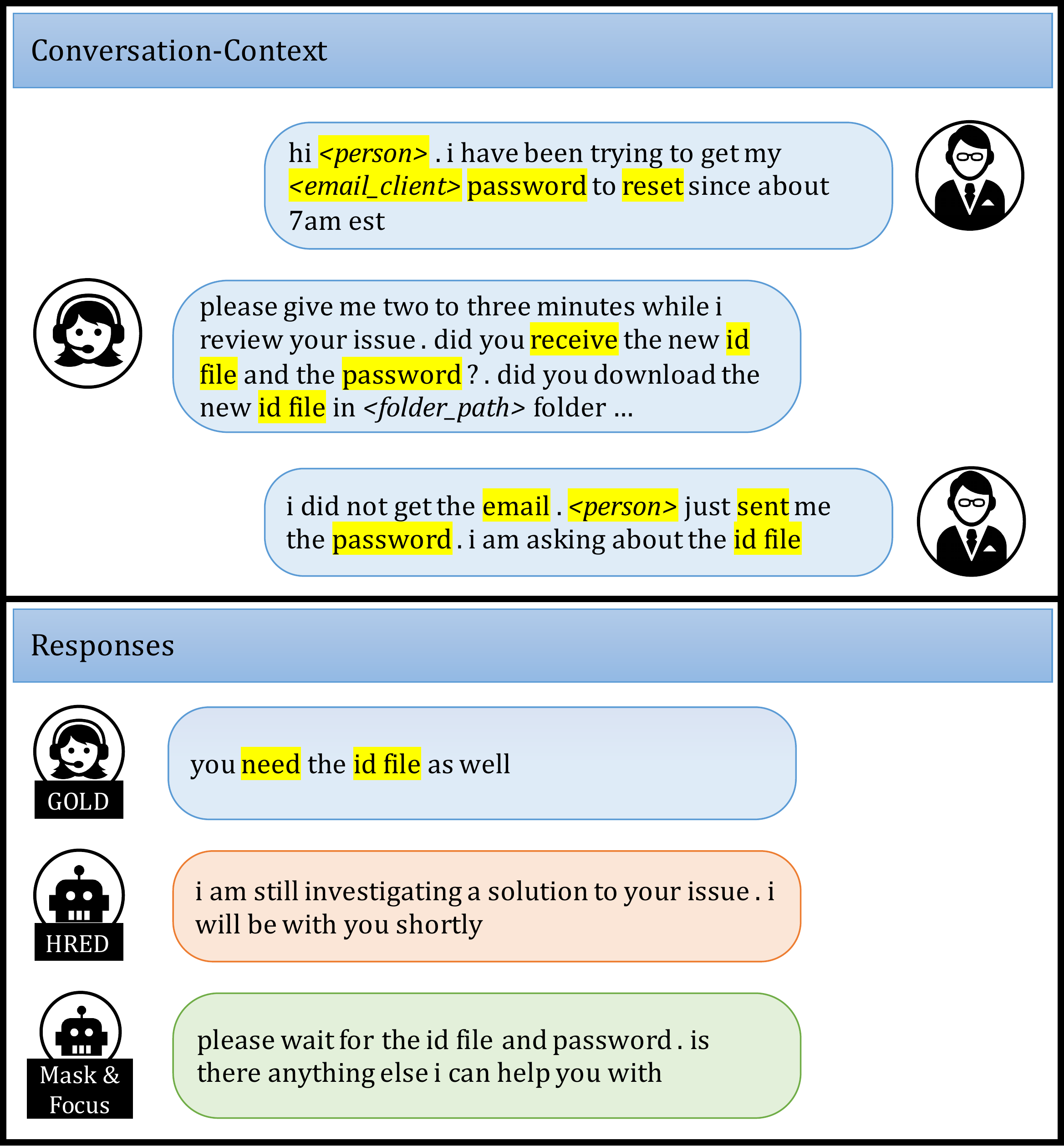}
    \caption{An example conversation from a tech support dataset along with responses generated responses by HRED and Mask \& Focus. The highlight indicates concepts identified by Mask \& Focus.}
    \label{fig:intro}
\end{figure}
In contrast to these models, humans focus primarily on those parts of the context that are relevant to generate a response. Moreover, we form a rough sketch of the important words/phrases to be mentioned in the response, before letting the language model take over. This process is illustrated using an example conversation in Figure \ref{fig:intro}. The spans in the conversation context necessary to generate the response are highlighted. The spans that convey the crux of the response are also highlighted in the gold response. We refer to these important spans that captures the semantics of the conversations as concepts.
%In comparison to using all words in the context and response, using just the concept to learn the semantics will prove useful when the amount of data available is limited.

Past work has shown \cite{serban2017multiresolution} that representing contexts and responses using concepts improves the performance of conversation modeling but they require a tedious process of manual identification of these concepts for each domain. In this paper, we propose Mask \& Focus, an approach which mimics human response generation process by discovering concepts in an unsupervised fashion. The model maps the context to the concept space. The concepts and the context are then used to predict the response concepts. 
%The approach first discovers concepts in both contexts and responses. Then it learns to generate response concepts from context concepts. 
Finally, the response concepts, along with the context and context concepts are used to generate the response. Learning to converse through concept has three main advantages: (1) It disentangles the learning of language and semantics in conversation modelling. The context in natural language is mapped to concept or semantic space. The model then translates the context concepts to response concepts. Finally the response concepts are mapped back to natural language. (2) Using words/phrases to represent concepts in context and response makes them interpretable (3) When the amount of available data is limited, learning in the concept space can be much more data-efficient. Figure \ref{fig:intro} shows an example response generated by Mask \& Focus and HRED on a tech-support dataset. The example illustrates the ability of Mask \& Focus to leverage the identified concepts to generate a topically coherent response, while HRED resorts to a generic response. 

In order to discover concepts in an unsupervised fashion, we systematically probe a trained conversation model by \textit{mask}ing words in the context and computing the pointwise mutual information (PMI) of the masked words with respect to the response. Then, the decoder is trained to predict the response concepts by optimizing a variational lower bound~\cite{jordan1999introduction} to the marginal log-likelihood of the response. To ensure that the model focuses on the context concepts, we choose the variational distribution in the lower bound as a function of the mutual information between the concepts in the context and response. We show that Mask \& Focus outperforms existing approaches on two datasets on automatic evaluations and human evaluation study. %By modelling the concepts, we were able to ensure that the generated responses are informative for the context. Moreover, the model was able to learn several interesting concepts related to Ubuntu that weren't discovered by manual annotation or searching its package managers in~\cite{serban2017multiresolution}.

To summarize, we make the following contributions:
\begin{enumerate}
\item \emph{Mask}: We propose a novel technique for learning essential concepts in the conversation context by using PMI-guided systematic probing.
\item \emph{Focus}: We propose a novel response generation technique that focuses on the context concepts by optimizing a variational lower bound to the marginal log-likelihood of the response.
\item We show that Mask \& Focus achieves significant improvement in performance over baselines in several established metrics on Ubuntu dialog corpus and an internal technical support dataset.
\end{enumerate}

\section{Approach}

\begin{figure*}
    \centering
    \includegraphics[scale=0.5]{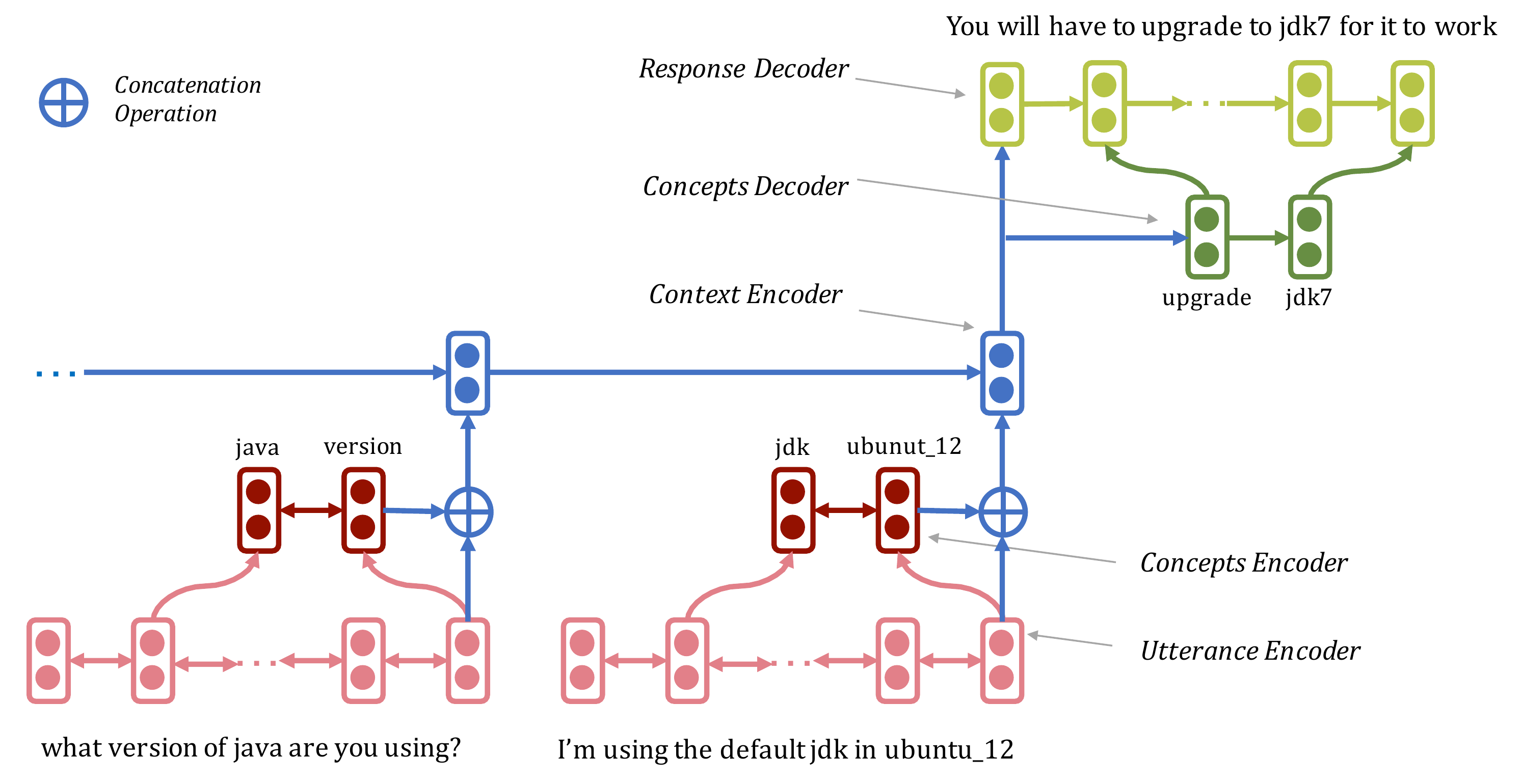}
    \caption{An overview of the proposed model. The bidirectional arrows indicate a bidirectional RNN while the unidirectional arrows indicate a unidirectional RNN. The curved arrows are used to indicate the words that are copied (either from utterances to concepts or vice-versa.) }
    \label{fig:mask_and_focus}
\end{figure*}{}

A conversation is a sequence of utterances $(u_1, \ldots, u_m)$ between two or more individuals. Moreover, an utterance $u_i$ is a sequence of words $(w_{i1}, \ldots, w_{il})$. At any given time-step $t<m$, the utterances till time $t$ are referred to as the context $\mathbf{c}$, whereas the $(t+1)^{th}$ utterance is referred to as the response $\mathbf{r}$. The set of all context response pairs in the training data is denoted by $\mathcal{D}$. The set of all words used for training a model is referred to as the vocabulary and is denoted by $\mathcal{V}$.

\subsection{Generative Conversation Modelling}
Generative conversation models maximize the probability of the groundtruth response given the input context. That is, for a given context response pair $(\mathbf{c},\mathbf{r})$ from the training data, they maximize the probability of the response $\mathbf{r}$ given the context $\mathbf{c}$. Towards that end, the context is fed to an encoder to get a context embedding. The encoder can either be a flat RNN or a hierarchical RNN for processing words and utterances at different levels. The context embedding is fed to a decoder and the probability of the groundtruth response is maximized. In other words, generative conversation models maximize the following quantity:
\begin{equation}
    \mathcal{L}(\theta) = \sum_{(\mathbf{c},\mathbf{r}) \in \mathcal{D}} \log p(\mathbf{r}|\mathbf{c} ; \theta)\,,
\end{equation}
where $\theta$ is the set of parameters that parametrize the above distribution. 
For a trained generative conversation model $\mathcal{G}$, the corresponding response distribution will be denoted as $p_{\mathcal{G}}(\mathbf{r}|\mathbf{c})$. 

\subsection{Mask \& Focus}

Given the context $\mathbf{c}$, we predict the context concepts $\mathbf{z}_c$. The context and the context concepts are used together to predict the response concepts $\mathbf{z}_r$. Finally, the context $\mathbf{c}$, and the concepts $\mathbf{z}_c$ and $\mathbf{z}_r$ are used for predicting the response. An overview of the model is shown in Figure~\ref{fig:mask_and_focus}.

\subsubsection{Mask-Context Concepts:}\label{sec:context}
While generative conversation models capture the correlation between the words in the context and response, they tend to generate generic and non-informative responses for most contexts. This is due to the fact that these models fail to identify the words in the context that need to be focused while generating a response. In contrast, humans selectively focus on the essential concepts discussed in the context and frame a response accordingly. 
We wish to mimic this mechanism by identifying the essential concepts in the context. We restrict our context concepts to be subsets of words present in the context.

Given a context-response pair $(\mathbf{c},\mathbf{r})$, the context concepts $\mathbf{z}_c$ are those words in the context that are `vital' for generating the response $\mathbf{r}$. The concepts are unknown and hence, they need to be learnt from the data. We learn these concepts by maximizing the pointwise mutual information between these concepts and the response.

For any given word $w$ in the vocabulary, the context obtained after masking the word is referred to as the masked context: $\mathbf{c}^{-w} = \mathbf{c}-w$. 
Conditioned on the masked context, the PMI between the word $w$ and the response $\mathbf{r}$ for a pretrained conversation model $p_{\mathcal{G}}$ is given by
\begin{align}
   \mathrm{PMI}(\mathbf{r}; w)  &= \log \frac{ p_{\mathcal{G}}(\mathbf{r}, w|\mathbf{c}^{-w})}{p_{\mathcal{G}}(\mathbf{r}|\mathbf{c}^{-w}) p_{\mathcal{G}}(w|\mathbf{c}^{-w}) } \notag \\
   &= \log  p_{\mathcal{G}}(\mathbf{r},| \mathbf{c}^{-w}, w) - \log p_{\mathcal{G}}(\mathbf{r}|\mathbf{c}^{-w}) \notag\\
   &= \log p_{\mathcal{G}}(\mathbf{r}| \mathbf{c}, w) - \log p_{\mathcal{G}}(\mathbf{r}|\mathbf{c}^{-w})
\end{align}
Here, the last line follows since $\{w\} \cup \mathbf{c}^{-w} = \mathbf{c}$. For the generative conversation model $p_\mathcal{G}(\mathbf{r}|w,\mathbf{c}) = p_\mathcal{G}(\mathbf{r}|\mathbf{c})$.
Hence, the pointwise mutual information between a word and the response is the difference between the log-probability of the response when the word $w$ is present and the log-probability of the response when $w$ is absent from the context. Intuitively, this quantity captures the importance of word $w$ for predicting the response.
This definition can easily be extended from a word to any set of words. 

Hence, we propose the following strategy for learning the context concepts: A context-response pair is sampled from the data. The context is probed systematically to obtain a masked context. The probe randomly masks the words in the context.
The context as well as the masked context are passed through a generative conversation model to evaluate the log-probability of the response before and after masking. The difference in log-probabilities are used for updating the scores of the masked words. The words with scores above a certain threshold are pushed to the concept bank. The context concepts $\mathbf{z}_c$ are all those words in the context that are also available in the concept bank. 

Having obtained the context concepts $\mathbf{z}_c$, we encode them as follow:
Each utterance in the context is fed to an RNN encoder to obtain the utterance embedding. The corresponding concepts are also fed as input to an RNN encoder and the resultant embedding is concatenated with the utterance embedding. The embeddings of all utterances and corresponding concepts are fed to a second-level RNN to generate the context vector. This context vector will be denoted as $\mathbf{v}$. This corresponds to the lower-left portion of Figure~\ref{fig:mask_and_focus}.

\subsubsection{Focus-Response Concepts:}
Since neural conversation models are black boxes, the concepts learnt by probing a conversation model can be useful for obtaining a better understanding of the trained model. However, our aim here is utilize the concepts learnt so far for generating better responses. Specifically, we want the model to generate responses that take into account the context concepts. 

An obvious strategy is to provide the context concepts $\mathbf{z}_c$ learnt in the previous section along with the corresponding utterances as input to the encoder as discussed in the end of the previous section. Unfortunately, a model can still choose to ignore $\mathbf{z}_c$ while generating a response. Hence, instead of predicting the response directly from the context and context concepts, we break the process of response generation by predicting the response concepts $\mathbf{z}_r$ at an intermediate step. By choosing $\mathbf{z}_r$ to have high mutual information with $\mathbf{z}_c$, the model will be forced to focus on $\mathbf{z}_c$ and hence generate more informative responses.  

Given the context and it concepts, the joint distribution of the response and its concepts can be written as
\begin{align}
    p(\mathbf{r}, \mathbf{z}_r|\mathbf{c}, \mathbf{z}_c) = p(\mathbf{z}_r|\mathbf{c}, \mathbf{z}_c)  p(\mathbf{r}| \mathbf{z}_r, \mathbf{c}, \mathbf{z}_c)
\end{align}
Since the response concepts are unknown, it is not possible to directly optimize the above probability. Instead, we propose to optimize a lower bound to the marginal likelihood of the response:
\begin{equation} \label{eq:ELBO}
    \mathbb{E}_{q(\mathbf{z}_r)}\log \frac{p(\mathbf{r}, \mathbf{z}_r|\mathbf{c}, \mathbf{z}_c)}{q(\mathbf{z}_r)} \le \log\sum_{\mathbf{z}_r} p(\mathbf{r}, \mathbf{z}_r|\mathbf{c}, \mathbf{z}_c)
\end{equation}
The above inequality is a straightforward consequence of Jensen's inequality.
The term to the left of the inequality is referred to as the evidence lower bound objective~(ELBO). Interestingly, the inequality holds true irrespective of the choice of the distribution $q$, also known as the variational distribution. 

Standard variational training proceeds by optimizing ELBO with respect to the variational distribution $q$ and the model distribution $p$ alternatively. However, we want to capture the property that the response concepts have high mutual information with respect to the context concepts. Thus, to ensure this property, we optimize the above objective with respect to the model distribution only. The distribution $q$ is chosen to give high probability to those words in the response that have high mutual information with respect to the context concepts.

%As mentioned earlier, we want the response concepts to have high mutual information with respect to the context concepts. We capture this property by designing the variational distribution to capture the pointwise mutual information between the response concepts and the context concepts.
%The context $\mathbf{c}$ is probed using the , we  the masked context $\mathbf{c}^{-z}$ is obtained by masking all words in the context that are also contained in $\mathbf{z}_c$. 
Specifically, let $w_\ell$ be the $\ell^{th}$ word in the response. The pointwise mutual information (PMI) of $w_\ell$ with the context concepts $\mathbf{z}_c$  conditioned on the masked context and the previous words for the model $p$ is given by
\begin{align}
    \mathrm{PMI}(w_\ell, \mathbf{z}_c) &= \log \frac{p(w_\ell, \mathbf{z}_c|  \mathbf{c}^{-z})}{p(w_\ell| \mathbf{c}^{-z}) p(\mathbf{z}_c| \mathbf{c}^{-z})} \notag\\
    &= \log p(w_\ell| \mathbf{c}^{-z}, \mathbf{z}_c) - \log p(w_\ell|  \mathbf{c}^{-z}) \notag\\
    &= \log p(w_\ell|\mathbf{c}, \mathbf{z}_c) - \log p(w_\ell|  \mathbf{c}^{-z})
\end{align}
Here, the last line follows because $\mathbf{c} = \mathbf{c}^{-z} \cup \mathbf{z}_c$. 
Note that unlike the pretrained conversation model $p_{\mathcal{G}}$, the proposed model $p$ takes the context concepts explicitly as input as shown in Figure~\ref{fig:mask_and_focus}.
For the sake of brevity, we have not explicitly stated the dependence of $w_{\ell}$ on $w_{1:\ell-1}$ explicitly. 

To capture this PMI in the ELBO in~\eqref{eq:ELBO}, we define the variational distribution $q$ as follows:
\begin{equation}
    q(w_{\ell}) = \begin{cases} \frac{\mathrm{PMI}(w_{\ell}; \mathbf{z}_c)}{1+\mathrm{PMI}(w_{\ell}; \mathbf{z}_c)}, & \text{if } \mathrm{PMI}(w_\ell) >0 \\ 
    0 & \text{otherwise}
    \end{cases}
\end{equation}
Conditioned on the response, we assume that the response concepts are selected independently. Hence, the distribution of response concepts $q(\mathbf{z}_r)$ factorizes over the words in $\mathbf{z}_r$.
%The distribution of the response concepts $q(\mathbf{r}_z)$ is the product of the distribution of all words in $\mathbf{r}_z$. 

Next, we maximize the ELBO with respect to the model distribution $p$. Since $q$ is fixed for this step, the ELBO objective can be simplified as given below:
\begin{equation} \label{eq:ELBO2}
    \mathbb{E}_{q(\mathbf{z}_r)} \log p(\mathbf{z}_r|\mathbf{c}, \mathbf{z}_c) + \mathbb{E}_{q(\mathbf{z}_r)} \log p(\mathbf{r}|\mathbf{z}_r, \mathbf{c}, \mathbf{z}_c)
\end{equation}
The two terms in the above equation correspond to the training objective for the concept and response decoder respectively. We assume that the distribution of the response concepts is modelled autoregressively, that is:
\begin{equation}
    \log p(\mathbf{z}_r|\mathbf{c}, \mathbf{z}_c) = \sum_{w_{\ell} \in \mathbf{z}_r} \log p(w_\ell|w_{1:\ell-1}, \mathbf{c}, \mathbf{z}_c)
\end{equation}
Hence, the first term in~\eqref{eq:ELBO2} can be approximated by expanding it over the words in the response as follows:
\begin{equation} \notag
    \mathbb{E}_{q(\mathbf{z}_r)} \log p(\mathbf{z}_r|\mathbf{c}, \mathbf{z}_c) \approx \sum_{w_{\ell} \in \mathbf{r}} q(w_\ell)\log p(w_\ell|w_{1:\ell-1}, \mathbf{c}, \mathbf{z}_c)
\end{equation}
The distribution $p(w_\ell|w_{1:\ell-1}, \mathbf{c}, \mathbf{z}_c)$ is implemented using a decoder RNN. The variational distribution $q$ weighs the words of the response for training the concept decoder.

% Assuming the domain of $\mathbf{z}_r$ to be all the words in the vocabulary, the expectation in the first term can be expanded as follows:
% \begin{equation} \notag
%     \mathbb{E}_{q(\mathbf{z}_r)} \log p(\mathbf{z}_r|\mathbf{c}, \mathbf{z}_c) = \sum_{w_{\ell} \in \mathbf{r}} q(w_\ell)\log p(w_\ell|w_{1:\ell-1}, \mathbf{c}, \mathbf{z}_c)
% \end{equation}
% The distribution $p(w_\ell|w_{1:\ell-1}, \mathbf{c}, \mathbf{z}_c)$ is implemented using a decoder RNN. The variational distribution $q$ weighs the words of the response for training the concept decoder.

The second term in \eqref{eq:ELBO2} corresponds to the training objective for the response decoder. To optimize this term, we generate samples from the variational distribution $\mathbf{z}_r \sim q(\mathbf{z}_r)$. A decoder with copy mechanism~\cite{gu2016incorporating} is trained to predict the words of the response by copying the response concepts. 

Note that changing the model distribution $p$ for training the concept and response decoder causes the variational distribution $q$ to change. Hence, the computation of $q$ needs to alternate with training of concept and response decoders.

\section{Experimental Setup}
To demonstrate the utility of Mask \& Focus, we evaluate it on two datasets.
\subsubsection{Ubuntu Dataset:}
The proposed model is evaluated for generating responses on the Ubuntu Dialogue Corpus~\cite{lowe2015ubuntu}. This dataset has been extracted from Ubuntu chatlogs, where users seek support for technical problems related to Ubuntu. The conversations in this dataset deal with a wide variety of technical problems related to Ubuntu OS and hence the total number of concepts discussed in these conversations is indeed quite large. This makes this dataset an ideal candidate for evaluating the proposed dialogue model. 
%{\bf should we also mention about manually known set of concepts for this dataset here?}
Table~\ref{tab:dataset} depicts some statistics for this dataset:

\begin{table}[h]
\centering
\begin{tabular}{l r r}
\toprule
& \textbf{Ubuntu} & \textbf{Tech Support} \\ \midrule
Training Pairs & 499,873 & 20,000\\
Validation Pairs & 19,560 & 10,000\\
Test Pairs & 18,920 & 10,000\\
Average No. of Turns & 8 & 11.9\\
Vocabulary Size & 538,328 & 293,494\\
\bottomrule
\end{tabular}
\caption{Dataset statistics for Ubuntu Dialog Corpus v2.0 ~\cite{lowe2015ubuntu} and Tech Support dataset}
\label{tab:dataset}
\end{table}

\begin{comment}
\begin{table}[h]
\centering
\begin{tabular}{lr}
\hline
Training Convs & 20,000 \\
Validation Convs & 5,000\\
Test Convs & 10,860\\
Average No. of Turns & 11.9 \\
\\
$|V|$ & 293,494\\
\hline
\end{tabular}
\caption{Dataset statistics for Tech Support dataset.}
\label{tab:tech-dataset}
\end{table}
\end{comment}
\subsubsection{Tech Support Dataset:}
We also conduct our experiments on an internal technical support dataset with $\sim20$K conversations.  We will refer to this dataset as Tech Support dataset in the rest of the paper. Tech Support dataset contains conversations pertaining to an employee seeking assistance from a technical support agent to resolve problems such as password reset, software installation/licensing, and wireless access. In contrast to Ubuntu dataset, this dataset has clearly two distinct users --- employee and agent. In our experiments we model the \textit{agent} responses only.
For evaluating the efficacy of Mask \& Focus for low resource settings, we create subsets of this dataset with $5,000, 10,000$ and $20,000$ conversations respectively.
%Note that multiple context-response pairs can be generated from a single conversation. For each conversation, we sample 25\% of the possible context-response pairs. We create validation pairs by selecting $5000$ conversations randomly and sampling context response pairs). Similarly, we create test pairs from a different subset of $5000$ conversations. The remaining conversations are used to create training context-response pairs. Table~\ref{tab:tech-dataset} depicts some statistics for this dataset:

\begin{table*}[ht]
\centering
\begin{tabular}{lllllllll}
& \multicolumn{3}{c}{\bf{Entity}} & \multicolumn{3}{c}{\bf{Activity}} & \bf{Tense} & \bf{Cmd}\\
\toprule
\bf{Model} & \bf{F1} & \bf{P} & \bf{R} &  \bf{F1} & \bf{P} & \bf{R} & \bf{Acc.} & \bf{Acc.}\\
\toprule
VHRED & 2.53 &	3.28 &	2.41 &	4.63 & \bf{6.43} &	4.31 &	20.2 & 92.02\\
HRED & 2.22 & 2.81 & 2.16 &	4.34 & 5.93 &	4.05 & 22.2 & 92.58\\
HRED-attend & 2.44 & 3.13 & 2.08 & 4.86 & 6.04 & 4.65 & \bf{29.75} & \bf{99.89}\\
MrRNN-Noun & 6.31 & 8.68 & 5.55 & 4.04 & 5.81 & 3.56 & 24.03 & 90.66\\
\bf{Mask \& Focus}  & \bf{7.82} & \bf{8.78} &	\bf{8.69} & \bf{5.43} & \bf{6.42} & \bf{5.90} &	24.47 &	93.19\\
\midrule
MrRNN-ActEnt & 3.72 & 4.91 &	3.36 &	11.43 & 16.84	&  9.72	& 29.01 &	95.04\\
\bottomrule
\end{tabular}
\caption{Activity \& Entity metrics for the Ubuntu corpus.  HRED, VHRED, MrRNN-Noun and MrRNN-ActEnt as reported by Serban et al.~\shortcite{serban2017multiresolution}. Note that MrRNN-ActEnt is supervised since the activities and entities used in the evaluation metrics were also used during training.}
\label{tab:results_ubuntu}
\end{table*}

\begin{comment}
\begin{table*}[ht]
\centering
\begin{tabular}{llllllllll}
& \multicolumn{3}{c}{\bf{Activity}} & \multicolumn{3}{c}{\bf{Entity}} & \bf{Tense} & \bf{Cmd} & \bf{BLEU}\\
\toprule
\bf{Model} & \bf{P} & \bf{R} & \bf{F1} & \bf{P} & \bf{R} & \bf{F1} & \bf{Acc.} & \bf{Acc.} & $\times$ 100\\
\toprule
VHRED* & 6.43 &	4.31 &	4.63 &	3.28 &	2.41 &	2.53 &	20.2 & 92.02  & 0.51\\
HRED* & 5.93 &	4.05 &	4.34 &	2.81 &	2.16 &	2.22 &	22.2 & 92.58 & 0.65\\
MrRNN-Noun* & 5.81 & 3.56 & 4.04 & 8.68 & 5.55 & 6.31 & 24.03 & 90.66 & 0.17\\
\\
\bf{Mask \& Focus}  &  \bf{8.88}    &   \bf{6.8} &	\bf{6.92} &  \bf{9.44} &	\bf{8.76} &	\bf{8.15} &	\bf{24.31} &	\bf{94.63} & \bf{0.70} \\
\midrule
MrRNN-ActEnt & \bf{16.84}	& \bf{9.72}	& \bf{11.43} & 4.91 &	3.36 &	3.72 & 29.01 &	95.04  & 0.17\\
\bottomrule
\end{tabular}
\caption{Activity \& Entity metrics for the Ubuntu corpus.  HRED*, VHRED*, MrRNN-Noun and MrRNN-ActEnt as reported by Serban et al.~\shortcite{serban2017multiresolution}. Note that MrRNN-ActEnt is supervised since the activities and entities used for evaluating the model were also used during training.}
\label{tab:results_ubuntu}
\end{table*}
\end{comment}

\subsection{Baselines}
We compare the performance of the proposed model against several models that have been proposed for modelling conversations in the past few years: $(i)$ HRED: An extension of sequence to sequence models for dealing with sequence of sequences~\cite{serban2016building} $(ii)$ VHRED: An extension of HRED to incorporate continuous-valued latent variables~\cite{serban2017hierarchical}. This model is trained using variational inference with reparametrization~\cite{diederik2014auto}.
$(ii)$ HRED-attend: An extension of HRED that attend on the words of the context.
$(iii)$ MrRNN-Noun: A multi-resolution conversation model where the nouns from the context and response are identified and modelled in conjunction with the actual context and response~\cite{serban2017multiresolution}.  $(iv)$ MrRNN-ActEnt: A multi-resolution conversation model where the activities and entities are predefined for the conversations. MrRNN-ActEnt models the relationship between these predefined activities and entities via an RNN in conjunction with the actual context and response.

\subsection{Implementation Details}
We implemented Mask \& Focus using the Pytorch library~\cite{paszke2017automatic}. The word embeddings as well as weights of various neural networks are initialized using a standard normal distribution with mean $0$ and variance $0.01$. The biases are initialized with $0$. We use $500$-dimensional word embeddings for all our experiments. 

For the Ubuntu dataset, the utterance and utterance concept encoders are single-layer bidirectional encoders, where each direction has a hidden size of $1,000$. The context encoder is a single-layer unidirectional encoder with a hidden size of $2,000$. We use a single-layer decoder with a hidden size of $2000$ for decoding the response concepts. The response decoder with copy attention also has a hidden size of $2,000$. We use a fixed vocabulary size of $20,000$. The word embeddings are shared by the utterance and utterance concept encoders. Similarly, the response concept decoder as well as the response decoder also share the word embeddings.
A similar model is also used for Tech Support dataset, with the exception that the dimensions of hidden layers are halved for both encoders and decoders.

We used the Adam optimizer with a learning rate of $1.5 \times 10^{-4}$. A batch size of $10$ conversations is used for training. Note that all the responses in the conversation are trained simultaneously by treating all utterances preceding the response as the context. As discussed before, an HRED is pretrained and is used for guiding the training of context concepts. The dimensions of utterance and context encoder as well as the response decoder are exactly the same as their corresponding dimensions in the proposed model. To prevent the model from overfitting, we use early stopping with log-likelihood on validation set as evaluation criteria.

\section{Results}

\subsubsection{Quantitative Evaluation:}
The metrics that are often used for comparing machine translation models have been shown to be uncorrelated with human judgements for conversation modelling~\cite{liu2016not}. This is understandable since a single context can be associated with multiple correct responses that have little word-overlap with each other. 

Hence, a new set of metrics were proposed in~\cite{serban2017multiresolution}, which compare responses based on a list of predefined entities and activities. The coarse-level representation of the predicted response is obtained in terms of activities and entities and compared with the corresponding representation of the ground truth response. These metrics were designed specifically for the Ubuntu Dialogue Corpus. A brief description of each metric is given below:
\begin{itemize}
\item \textbf{Activity}:  Activity metric compares the activities present in a predicted response with the ground truth response. Activity can be thought of as a verb.
\item \textbf{Entity}: This compares the technical entities that overlap with the ground truth response. A total of 3115 technical entities were identified using public resources such as Debian package manager APT.
\item \textbf{Tense}: This measure compares the tense of ground truth with predicted response.
\item \textbf{Cmd}: This metric computes accuracy by comparing commands identified in ground truth utterance with a predicted response.
\end{itemize}

We use the code provided by~\cite{serban2017multiresolution} for computing precision, recall and F1-score for entities and activities.
Table~\ref{tab:results_ubuntu} compares the proposed model against several baselines based on each of the above metrics. MrRNN-ActEnt learns the activities and entities in a supervised manner, that is, the information about activities and entities is available during training. In contrast, all the other models used for comparison are trained in an unsupervised manner. As can be observed from the table, the proposed model significantly outperforms the unsupervised baselines on each metric. This suggests that Mask \& Focus can predict the correct activities and entities with higher accuracy and hence, can aid in resolving the technical support problems of Ubuntu dataset.

Even more surprisingly, we were able to significantly outperform the supervised baseline MrRNN-ActEnt for the entity metrics. 
This indicates that the focus-based approach for learning response concepts is better than directly identifying the entities and training a supervised model to predict them. Moreover, we were able to discover several interesting concepts that were not identified manually in 
MrRNN-ActEnt. Some of these concepts are shown in Table~\ref{tab:ubuntu_con}. 

\begin{table}
\centering
\small
\begin{tabular}{|p{0.95\linewidth}|}
\hline
trash, glxgears, DNS, fonts, ndiswrapper, clamscan, vmware, torrent, inittab, gmake, libmp3lame, sound, Beryl, cube, firewall, bogomips, ipv6, dragon, updates, smbfs, envy, raid, antivirus, suspend, madwifi, fakeraid,  runlevel, workspaces, tar, mono, Define, burning, netstat, proxy, automatix, port, NickServ, RAID, font, chkconfig, truecrypt, fsck, apache, realplayer, beryl, burn, initrd, evolution, dns, xampp, SD, skydome, brightness, photoshop, muine, patch, mbox, locale, logs, Firewall, mic, 64-bit, flash64, game, dd, spanish, dock, lsb\_release, webmin, xvid, transparency, JDK, /boot, filesharing, motd, dist-upgrade, netflix, vi, encrypt, /etc/environment, wubi, pcspkr, IA64, OpenOffice, serpentine, i686, mkinitrd, upgrade, theme, uname, Apache, ltsp, %visudo, XAMPP, kad, bootsplash, hardy-proposed, bash-completion, nessus, clock, programming, chrome, multitouch
\\
\hline
\end{tabular}
\caption{The top scoring concepts in Ubuntu that weren't manually annotated by ~\cite{serban2017multiresolution} for training MrRNN-ActEnt.}
\label{tab:ubuntu_con}
\end{table}

%To verify this hypothesis, we replaced the focus based approach presented in this paper and set the weight of each response concept to be the same. In particular, for each word in the response, we compute the difference between the loss before and after masking of context concepts. If this difference is greater than a certain threshold, the word is identified as a response concept. Each response concept is assigned the same weight, and a decoder is trained directly to predict the identified concepts. We compare the performance of this modified model with respect to the original model on entity metrics. As can be observed from the table, entity scores drop significantly.

% \begin{table}[h]
% \centering
% \begin{tabular}{lrr}
% \toprule
% \bf{Entity} & \bf{Mask} & \bf{Mask}\\
% \bf{scores} & \bf{and Focus} & \bf{w/o Focus}\\

% \midrule
% Precision & 9.44 & 7.11\\
% Recall & 8.76 & 5.68\\
% F1 & 8.15 & 5.71\\
% \bottomrule
% \end{tabular}
% \caption{Effect of removing the Focus model for learning response concepts}
% \end{table}

\subsubsection{Low Resource Setting:} 
\begin{figure}
    \centering
    \includegraphics[scale=0.55]{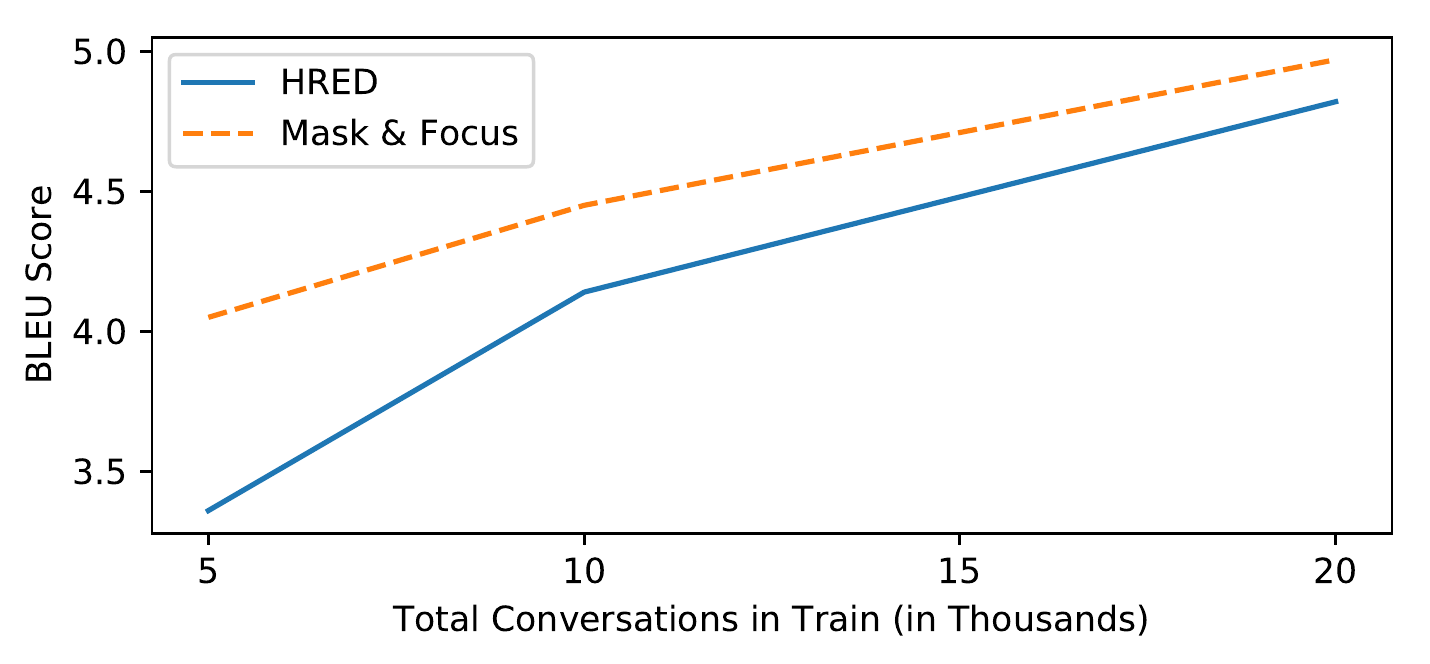}
    \caption{Comparison of Mask \& Focus against HRED for Technical Support dataset.}
    \label{fig:tech_comp}
\end{figure}

Next, we evaluate the performance of Mask \& Focus on the Technical Support Dataset. As the Technical Support dataset is from a narrower domain than the Ubuntu corpus, we use this dataset to understand if Mask \& Focus is able to capture these concepts even when the amount of resources available is small. In order to show this, we create subsets with randomly sampled 5,000, 10,000 and 20,000 context response pairs from the dataset and train HRED and Mask \& Focus on these subsets. Unlike Ubuntu, we do not have a predefined list of activities and entities for this dataset. Hence, we use BLEU score~\cite{papineni2002bleu} to comapre the performance. The results are shown in Figure~\ref{fig:tech_comp}. As can be observed from the Figure, Mask \& Focus outperforms HRED considerably when the dataset size is small. However, as the dataset grows in terms of training samples the gap reduces.

\subsubsection{Human Evaluation}
\citeauthor{liu2016not}~\shortcite{liu2016not} have shown that metric based evaluations of response prediction in conversations are not strongly correlated with human judgements. So, we conduct a human study to compare the quality of the responses generated by Mask \& Focus and other existing approaches.  Annotators were shown a conversation-context along with the responses generated by unsupervised models (i.e) Mask \& Focus, HRED, VHRED and MrRNN-Noun. The name of the algorithms were anonymized and we randomly shuffled the order as well. All annotators were asked to rate the relevance of the response on a scale of 0-4. We noticed that existing approaches often generated generic responses such as "what are you talking about" or "i see". So, we also requested the annotators to mark such generic responses. We removed these generic responses before computing the average relevance score reported in Table \ref{tab:human_judgements}. We randomly sampled 100 context response pairs from the dataset, and collected a total of 500 relevance annotations. 

\begin{table}[h]
\centering
\begin{tabular}{lcc}
\toprule
\textbf{Model} & \textbf{Relevance} & \textbf{GR Ratio}\\ 
\midrule
Gold &  2.66 & 0.05\\ \midrule
HRED &  1.25 & 0.53\\
VHRED &  1.48 & 0.39\\
MrRNN-Noun &  1.56 & 0.32\\
Mask \& Focus &  \textbf{1.79} & \textbf{0.18}\\
\bottomrule
\end{tabular}
\caption{Human judgements on model responses on 100 random samples from Ubuntu corpus. Relevance scores are on a scale of (0-4). GR stands for Generic Response}
\label{tab:human_judgements}
\end{table}

Table \ref{tab:human_judgements} shows the relevance scores of gold and various algorithms. GR Ratio indicate the ratio of responses that were generic. Mask \& Focus's ability to produce topically coherent responses is reflected by the significantly low generic response ratio. The relevance scored of Mask \& Focus is significantly better that the existing approaches. A low relevance score on the gold responses captures the inherent difficulty in modelling the Ubuntu Corpus.

\begin{table*}[h]
\centering
\begin{tabular}{|l|l|l|l|l|l|l|l|}
\toprule
ipod & unrar & webcam & skype & eclipse & pdf & java & raid \\
\midrule
ipod & unrar & webcam & skype & eclipse & pdf & java & raid \\
gtkpod & rar & cheese & mic & ide & evince & sun & software\\
itunes & unrar-nonfree & cam & deb & jdk & adobe & jre & mdadm  \\ 
ipods & multiverse & webcams & voice & netbeans & print & sun-java6-plugin  & array \\
rockbox & unrar-free & camera & medibuntu & sdk & xpdf & eclipse & controller  \\
/media/ipod & .rar & camorama & call & sun & reader & sun-java6-jre & hardware  \\
amarok & file-roller & lsusb & ekiga & pycharm & document & plugin & raid1 \\
banshee & sources.list & logitech & libqt3c102-mt & jre & pdfs & openjdk & fakeraid 
\\
rhythmbox & non-free & motion & sip & java6 & acrobat & update-alternatives & dmraid \\ 
plug & nonfree & amsn & uvcvideo & javac & openoffice & gcj & fake \\
\bottomrule
\end{tabular}
\caption{Context concepts (top row) and their corresponding response concepts. }
\label{tab:res_con}
\end{table*}

\begin{table*}[h]
\centering
\begin{tabular}{p{7cm} p{9.5cm}}
\toprule
\textbf{Conversation-Context} & \textbf{Responses} \\
\midrule 
can someone tell me \textbf{how} to install the \textbf{downloads} \textbf{i} got off the \textbf{internet} \textbf{?} $\rightarrow \cdots \rightarrow$ if you need help otherwise you need to name the \textbf{package} type you have like is it a \textbf{tar}.gz etc $\rightarrow$ its a .\textbf{rar} and a .\textbf{exe} & \textbf{mask \& focus:} you need to \textbf{unrar} the package \textbf{manager} to \textbf{unrar} \textbf{rar} \textbf{archives} , then it doesnt work with the \textbf{archive} \textbf{manager} \newline  \textbf{hred:} What version of Ubuntu are you using ? \newline \textbf{vhred:} I think you need to install the . tar . gz . \newline \textbf{mrrnn-noun:} unk
\\ \midrule
my \textbf{java} has been faulty lately and \textbf{i} want to \textbf{reinstall} $\cdots$ could someone \textbf{please} help me \textbf{remove} all my \textbf{java} \textbf{?} $\rightarrow$ and tell us which of the javas you have $\rightarrow$ \textbf{i} also have some downloaded from oracle & \textbf{mask \& focus:} \textbf{java} -version? what does \textbf{java} -version tell you? \newline  \textbf{hred:} what is the error? \newline \textbf{vhred:} and what is the output of : dpkg -l $\vert$ grep unk \newline \textbf{mrrnn-noun:} how do i uninstall oracle ?  \\ \midrule
you can use \textbf{gparted} to \textbf{resize} the win7 \textbf{partition} to make room for a \textbf{ubuntu} \textbf{partition} $\rightarrow$ \textbf{thanks} & \textbf{mask \& focus:} you can \textbf{resize} \textbf{ntfs} partitions , but \textbf{gparted} wont let you \textbf{resize} partitions \newline  \textbf{hred:} what is the error? \newline \textbf{vhred:} what version of ubuntu are you running ? \newline \textbf{mrrnn-noun:}is it possible to resize a partition ? \\
\bottomrule
\end{tabular}
\caption{Model Responses from Ubuntu Dialog Corpus. The context concepts and the response concepts are in bold. Arrow ($\rightarrow$) indicates end of turn.}
\label{tab:examples}
\end{table*}

\subsection{Discussion}
We inspect the response concepts associated with the concepts in the context in Table~\ref{tab:res_con}. To obtain this table, we evaluate Mask \& Focus on all conversations containing the specific context concept. The response concepts are predicted by the concept decoder. We list the most frequent response concepts in Table~\ref{tab:res_con}.
As can be observed from the Table, the response concepts learnt by the model are indeed quite meaningful. 
Moreover, masked probing of generative conversation models, discovers concepts that weren't even extracted by manual inspection of Ubuntu chatlogs or by extracting terms from the package managers (the approach used in~\cite{serban2017multiresolution}). %This suggests that the proposed approach of probing with masking is indeed quite useful for discovering interesting concepts from the context.

Responses generated by various algorithms are shown in Table \ref{tab:examples}. The contexts are highlighted with the discovered context concepts and the Mask \& Focus responses are highlighted with the copied response concepts. We see that Mask \& Focus is able to capture important concepts and generate a topically coherent response. For example, it was able to learn that if the context contains "download" and "tar", then the response should have "unrar".

\section{Related Work}
The idea of generating an abstract representation of the response before generating the response has been explored by Serban et .al. \shortcite{serban2017hierarchical} and Zhao et. al. \shortcite{zhao2017learning}. \citeauthor{pandey2018exemplar}~\shortcite{pandey2018exemplar} proposed to ground the generated response in a set of retrieved responses. \citeauthor{liu2018knowledge}~\shortcite{liu2018knowledge} incorporated external knowledge to guide response generation in a dialog.
Serben et. al. \shortcite{serban2017multiresolution} proposed Multi resolution RNN (MrRNN) to augment sequence-to-sequence architecture to support multiple higher levels of abstractions of input and output sequences. However, unlike Mask \& Focus, the model requires the higher level abstractions to be pre-defined and available during training.  

%Unlike these approaches, Mask \& Focus uses abstraction for both the response and the context. In addition to that, the abstract representations are interpretable.

Systematic probing of trained models using manually defined masks has been explored by \citeauthor{hendricks2018women}~\shortcite{hendricks2018women} and \citeauthor{caglayan2019probing}~\shortcite{caglayan2019probing}  to overcome bias in captioning and multi-modal machine translation. Unlike these approaches, Mask \& Focus discovers masks required for probing in an unsupervised manner. Pointwise mutual information (PMI) has proven useful in conversation modelling for addressing the problem of generic responses~\cite{li2016diversity}. While \citeauthor{li2016diversity} used mutual information for reranking the candidate responses only, we incorporate it in our training mechanism for discovering context and response concepts. 

%Weighted cross entropy used by Mask \& Focus to emphasise the importance of copying response concepts to the response. Weighted cross entropy has been proven useful for spotting keywords in speech in low resource settings \cite{Panchapagesan2016MultiTaskLA} and introducing semantic diversity in text generation \cite{kovaleva2018similarity}.

\section{Conclusion}
In this paper, we presented Mask \& Focus, a conversation model that learns the concepts present in the context and utilizes them for predicting the response concepts and the final response. The concepts are learnt by a strategy that involves masking the concepts in the context and focusing on the masked concepts. We show that the proposed Mask \& Focus model discovers interesting concepts from challenging datasets that fail to be discovered by manual annotation. Mask \& Focus achieves significant improvement in performance over existing baselines for conversation modelling with respect to several metrics. Moreover, we also observe that Mask \& Focus generates interesting responses specific to the input context. 

\begin{comment}
\begin{figure}
    \centering
    \includegraphics[width=.8\linewidth]{ubuntu_new_con.png}
    \caption{The top scoring concepts in Ubuntu that weren't manually discovered in~\cite{serban2017multiresolution} for training MrRNN-ActEnt}
    \label{fig:ubuntu_con}
\end{figure}
\end{comment}

\bibliographystyle{aaai}
\bibliography{citations}
\end{document}